%% file: acl_latex.tex
\documentclass[11pt]{article}

\usepackage[preprint]{acl}

\usepackage{times}
\usepackage{latexsym}

\usepackage[T1]{fontenc}

\usepackage[utf8]{inputenc}

\usepackage{microtype}

\usepackage{inconsolata}

\usepackage{graphicx}

\usepackage[utf8]{inputenc} 
\usepackage[T1]{fontenc}    
\usepackage{url}            
\usepackage{amsfonts}       
\usepackage{nicefrac}       
\usepackage{xcolor}         

\usepackage{amsmath}
\usepackage{amssymb}
\usepackage{mathtools}
\usepackage{amsthm}

\usepackage{arydshln}
\usepackage{enumitem}
\usepackage{lipsum}
\usepackage{wrapfig, makecell}

\usepackage{bm}
\usepackage{tcolorbox}
\usepackage{xspace}
\usepackage{caption}

\usepackage{multirow}
\usepackage{microtype}
\usepackage{graphicx}
\usepackage{algorithm}
\usepackage{algorithmic}
\usepackage{subcaption}
\usepackage{booktabs} 

\newcommand{\alg}{\texttt{\textbf{HyFlash}}\xspace}

\title{HyFlash: A Prefix-Memory Prior for On-Device Speculative Decoding}

\author{Jongwoo Ko \qquad  \\
  Microsoft Applied Sciences Group \\
  \texttt{jongwooko@microsoft.com} \\
  \And
  Pashmina Cameron \\
  Microsoft Applied Sciences Group \\
  \texttt{pcameron@microsoft.com}}

\begin{document}
\maketitle

\input{tex/00_abstract}
\input{tex/01_introduction}
\input{tex/02_motivation}
\input{tex/03_method}
\input{tex/04_experiments}
\input{tex/05_conclusion}


\bibliography{custom}

\clearpage
\appendix
\input{tex/91_related}
\input{tex/94_hardware}
\input{tex/92_experimental_setup}
\input{tex/93_additional}

\end{document}

%% file: tex/00_abstract.tex
\begin{abstract}
Parallel block drafting accelerates speculative decoding by predicting multiple future positions in a single backbone pass, but on-device deployment constrains both drafter capacity and context storage. Under tight memory budgets, drafters must model long-range dependencies and fine-grained draft-position interactions without the growing KV-cache overhead of full-context attention. We propose \alg, a compact parallel drafter that introduces a prefix-memory prior to separate long-range context modeling from fine-grained draft-position interactions with limited capacity. Local attention captures recent context and draft-position interactions, while recurrent memory compresses historical target features into fixed-size states. This inductive bias enables effective local and long-range modeling with bounded draft-side memory. Experiments show that \alg achieves higher acceptance than full-context attention with substantially less KV storage and with larger gains for shallow drafters and long contexts.
\end{abstract}

%% file: tex/01_introduction.tex
\section{Introduction}

Speculative decoding accelerates autoregressive language model inference by using a lightweight drafter to propose tokens that a larger target model verifies in parallel~\citep{leviathan2023fast}. Recent parallel block drafters, such as DFlash~\citep{chen2026dflash} and DSpark~\citep{cheng2026dspark}, further reduce drafting latency by predicting multiple future positions in a single backbone pass. However, maintaining high draft acceptance requires effectively modeling both historical context and interactions among draft positions while keeping drafting overhead small.

This challenge becomes particularly pronounced in on-device deployment~\citep{microsoft2024surface, apple2024mini, nvidia2026orin}, where memory is severely constrained~\citep{liu2024mobilellm, xu2024edgellm}. Target weights, KV cache, and runtime overhead leave only a small residual budget for drafter parameters and context storage~\citep{tiwari2025quantspec}. Under limited model capacity, jointly modeling long-range dependencies and fine-grained intra-block interactions becomes difficult, even with full-context attention. Moreover, full attention incurs growing KV memory, while windowed attention restricts direct access to distant information~\citep{yang-etal-2026-longspec, yuan2026compact}. These limitations are especially relevant to long-horizon conversations and tool-using agents.

Existing studies address memory-efficient speculative decoding through different approaches. LongSpec~\citep{yang-etal-2026-longspec} bounds draft-side KV memory, while MASW~\citep{yuan2026compact} compresses historical KV representations. H-Spec~\citep{jiang2026hspec} combines recurrent drafting with target-KV reuse, whereas ReaDy~\citep{sun2026ready} leverages recurrent states from hybrid target models. These advances motivate a complementary question: \textit{how can a capacity-constrained parallel drafter effectively model both local interactions and long-range context without relying on additional depth or full-context attention?}


We propose \alg, a compact parallel drafter built around a prefix-memory prior for capacity-constrained drafting. \alg maintains persistent memory of committed target features while using local attention for recent context and bidirectional draft-position interactions. Rather than updating recurrent states through speculative positions, \alg lets them query a shared, read-only prefix memory. Our contributions are threefold:
\begin{itemize}[leftmargin=*, itemsep=0pt]
    \vspace{-3pt}
    \item We characterize the capacity--context tension in on-device parallel drafting, highlighting the challenges of modeling long-range dependencies with shallow drafters.
    \item We introduce a prefix-memory prior that separates historical context modeling from local and intra-block interactions, enabling effective drafting with bounded context memory.
    \item We demonstrate improved acceptance and decoding efficiency across two target models, with capacity scaling and targeted interventions validating the benefits of this inductive bias under limited model capacity.
\end{itemize}


%% file: tex/02_motivation.tex
\section{Memory Constraints in On-Device Parallel Drafting}
Parallel block drafters must accommodate their weights and context alongside the target model. On-device memory constraints therefore limit both drafter capacity and access to historical context.

\input{fig/motif}
\paragraph{Memory pressure constrains capacity.}
On-device speculative decoding must accommodate target weights, KV cache, and runtime overhead before allocating memory to the drafter. We consider an illustrative 12 GiB inference budget for 16 GB-class devices, reserving approximately 2.9 GiB for the operating system and other workloads (\autoref{app:hardware}). Using Phi-4-reasoning~\citep{abdin2025phi} as a geometry reference for our 14B-family prototype, the target KV cache alone requires 6.25 GiB at 32K context. Even with ideal 2-bit target weights, at most 0.84 GiB remains for the drafter and other inference overheads (\autoref{fig:motivation}a). Such tight budgets severely restrict drafter capacity, motivating architectures that make effective use of only a few layers.


\paragraph{Limited capacity challenges context modeling.}
Parallel drafting requires modeling both long-range dependencies and fine-grained interactions among draft positions, which becomes challenging under limited model capacity. Even with full-context attention, a compact drafter may struggle to effectively utilize historical information, while retaining the full KV cache incurs substantial memory overhead. Windowed attention reduces this cost but limits direct access to distant information, leading to substantial acceptance degradation as shown in Figure~\ref{fig:motivation}b. These observations highlight two complementary limitations: windowed attention restricts access to distant information, while full-context attention incurs growing memory costs without necessarily overcoming the modeling limitations of shallow drafters. These challenges motivate a prefix-memory prior that combines recurrent modeling of historical context with local attention for recent and intra-block interactions.

%% file: fig/motif.tex
\begin{figure}[t]
    \centering
    \includegraphics[width=1.0\columnwidth]{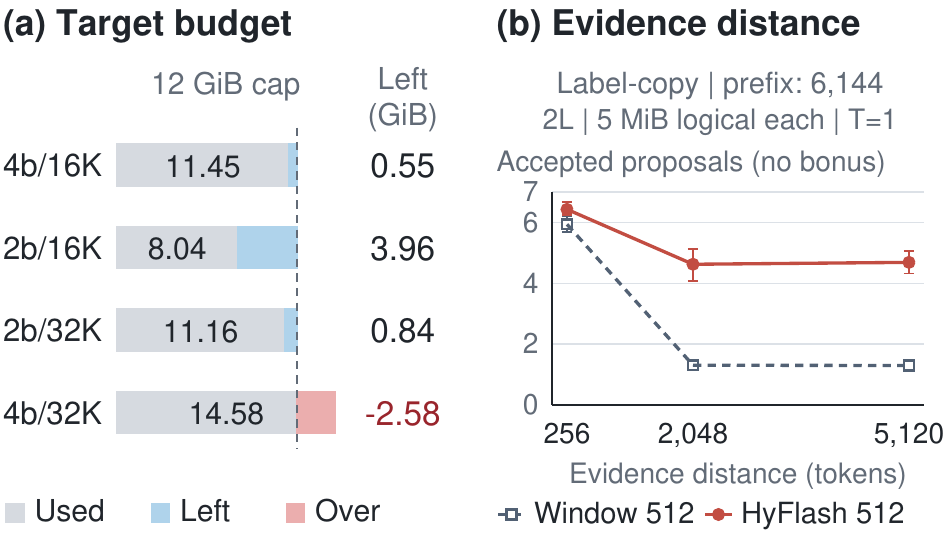}
    \caption{Motivation for memory-constrained drafting. (a) Analytical target memory usage under an illustrative 12 GiB inference budget for 16 GB-class devices (\autoref{app:hardware}). (b) Acceptance across evidence distances for independently trained two-layer drafters with equal 5 MiB logical context budgets.}
    \label{fig:motivation}
    \vspace{-10pt}
\end{figure}

%% file: tex/03_method.tex
\section{HyFlash}
We propose \alg, a compact parallel drafter combining local attention and recurrent memory within each layer to efficiently capture local interactions and long-range context with limited capacity.

\input{fig/memflash}
\vspace{-3pt}
\paragraph{Drafting from a committed prefix.} At boundary $a$, the target has processed $x_{<a}$ and selected a known anchor $x_a$, whose feature is not yet available. An adapter projects selected target hidden layers into $z_i$, $i<a$. Given these features, the anchor, and masked future positions, a single backbone pass produces representations for $K$ draft positions, which can be used with different proposal generation strategies.

\input{fig/main}
\vspace{-3pt}
\paragraph{Intra-layer hybrid attention.} Each layer implements the prefix-memory prior by combining sliding-window attention (SWA) for recent and draft-position interactions with recurrent memory that compresses historical target features. Unlike conventional hybrid sequence modeling \citep{dong2024hymba}, \alg explicitly
separates committed history from speculative positions: recurrent states are updated only from committed target features and remain read-only during parallel drafting. We instantiate the recurrent memory with Mamba-2~\citep{dao2024ssm}, whose selective state update follows
\begin{align}
    &S_i = \alpha_i S_{i-1} + (\Delta_i v_i) b_i^{\intercal}, \nonumber \\
    &\alpha_i = \exp (-\Delta_i \lambda), \quad \Delta_i,\lambda>0. \label{eq:state}
\end{align}
Here, $\Delta_i$ is an input-dependent step size, $\lambda$ is a learned head-specific decay rate, and $v_i$ and $b_i$ are bottleneck projections of committed target features. For each draft representation $u_k$, the local and recurrent branches retrieve recent and historical context, respectively. We define:
\begin{equation*}
    C_a=[z_i]_{i=\max(0,a-W)}^{a-1}\mathbin{\Vert}U,
\end{equation*}
where $U=[u_1,\ldots,u_K]$ denotes the draft-block representations. The two branches are computed and fused as
\begin{align}
A_k&=\operatorname{Attn}\!\bigl(Q(u_k),K(C_a),V(C_a)\bigr),\nonumber\\
r_k&=W_\uparrow
 \bigl[\operatorname{LN}(S_{a-1}^{h}c(u_k))\bigr]_{h=1}^{H_S},\nonumber\\
g_k&=\beta_A\odot\operatorname{RN}(A_k)
       +\beta_S\odot\operatorname{RN}(r_k).
\label{eq:fusion}
\end{align}
Local attention attends to the last $W$ committed target features and all draft-block positions bidirectionally, while the recurrent branch queries the compressed prefix state without modifying it. Here, $c$ is the shared read-query projection, $W_{\uparrow}$ projects concatenated head outputs, and $\beta_A, \beta_S$ are learned channel scales. LN and RN denote LayerNorm~\cite{ba2016layer} and RMSNorm~\cite{zhang2019root}, respectively. 

\paragraph{Persistent prefix memory.} 
Rather than relying solely on recent target features, \alg explicitly accumulates committed target features in a persistent recurrent state across drafting rounds. After verification, only newly processed, committed features update the memory, excluding unverified and rejected proposals.
Parallel draft positions query this memory without modifying it, preserving historical context without rescanning previous tokens.

\paragraph{Memory footprint.}
With bounded-window KV storage, \alg maintains a constant-size logical context footprint as the prefix grows. In our Phi-4-family configuration, local KV and FP32 recurrent states each require 2.5 MiB, together matching the 5 MiB logical context budget of a two-layer windowed drafter. This enables long-range context modeling without increasing drafter depth or the total logical context footprint. These estimates exclude model weights, target KV, and temporary buffers.

%% file: fig/memflash.tex
\begin{figure}[t]
    \centering
    \includegraphics[width=1.0\columnwidth]{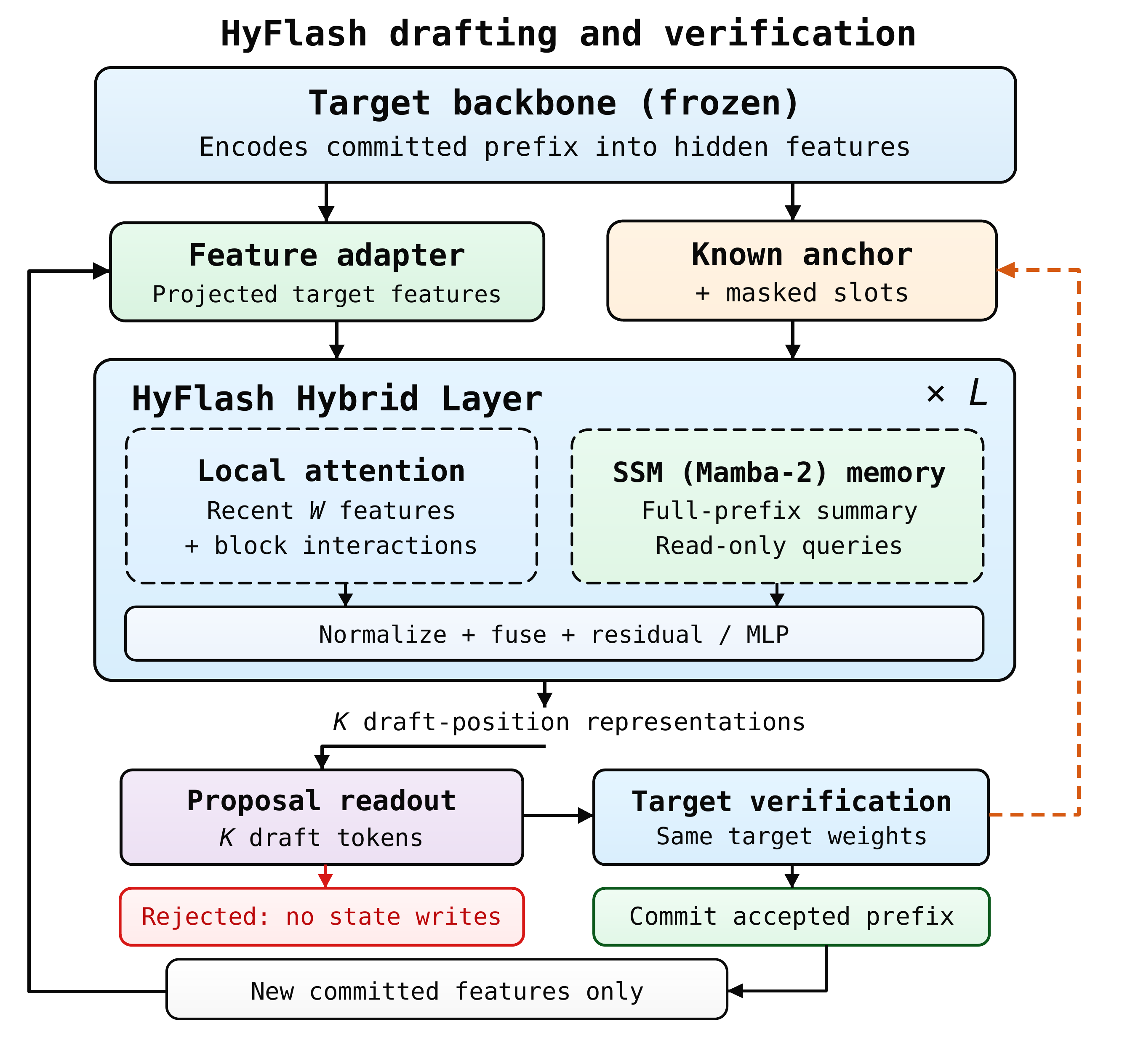}
    \caption{\alg framework. Target-conditioned hybrid layers produce representations for a separate proposal readout. The same target verifies proposals and returns the next anchor; only newly committed target features update memory.}
    \label{fig:memflash}
    \vspace{-10pt}
\end{figure}

%% file: fig/main.tex
\begin{table*}[t]
\centering
\setlength{\tabcolsep}{1.2pt}
\caption{Decoding speedup over autoregressive decoding and average acceptance length ($\tau$) across math, code, and chat benchmarks on Qwen3-4B and Phi-4-reasoning (14B). Both targets use thinking mode, with a maximum generation length of 4,096 tokens.}
\vspace{-5pt}
\resizebox{1.0\linewidth}{!}{
\begin{tabular}{@{}ll*{8}{cc}}
\toprule
\multirow{3}{*}{Target} & \multirow{3}{*}{Drafter}
 & \multicolumn{6}{c}{Math}
 & \multicolumn{6}{c}{Code}
 & \multicolumn{2}{c}{Chat}
 & & {}\\
\cmidrule(lr){3-8}\cmidrule(lr){9-14}
\cmidrule(lr){15-16}
 & & \multicolumn{2}{c}{GSM8K}
 & \multicolumn{2}{c}{MATH-500}
 & \multicolumn{2}{c}{AIME25}
 & \multicolumn{2}{c}{HumanEval}
 & \multicolumn{2}{c}{MBPP}
 & \multicolumn{2}{c}{LCB}
 & \multicolumn{2}{c}{MT-Bench}
 & \multicolumn{2}{c}{Avg.} \\
\cmidrule(lr){3-4}\cmidrule(lr){5-6}\cmidrule(lr){7-8}
\cmidrule(lr){9-10}\cmidrule(lr){11-12}\cmidrule(lr){13-14}
\cmidrule(lr){15-16}  
 &
 & Speedup & $\tau$ & Speedup & $\tau$ & Speedup & $\tau$
 & Speedup & $\tau$ & Speedup & $\tau$ & Speedup & $\tau$
 & Speedup & $\tau$ & Speedup & $\tau$ \\
\midrule
\multirow{3}{*}{Q3-4B} & DFlash
 & 1.89 & 3.19 & 1.80 & 3.28 & 1.64
 & 3.12 & 1.70 & 3.02 & 1.63 & 2.89
 & 1.36 & 2.75 & 1.53 & 2.63 & 1.64 & 2.98 \\
 & DSpark
 & 2.28 & 3.92 & 2.11 & 4.02 & 1.93
 & 3.82 & 2.04 & 3.73 & 1.91 & 3.54
 & 1.66 & 3.46 & 1.79 & 3.19 & 1.95 & 3.67 \\
 & \alg
 & \textbf{2.53} & \textbf{4.38} & \textbf{2.30} & \textbf{4.48} & \textbf{2.16}
 & \textbf{4.38} & \textbf{2.23} & \textbf{4.13} & \textbf{2.08} & \textbf{3.94}
 & \textbf{1.82} & \textbf{3.83} & \textbf{1.98} & \textbf{3.59} & \textbf{2.15} & \textbf{4.10} \\
\midrule
\multirow{4}{*}{P4-14B}
 & DFlash
 & 2.09 & 3.13 & 2.38 & 3.58 & 2.04
 & 3.10 & 2.21 & 3.26 & 2.20 & 3.17
 & 1.75 & 2.78 & 1.62 & 2.28 & 2.03 & 3.04 \\
 & DSpark
 & 2.39 & 3.48 & 2.51 & 3.79 & 2.14
 & 3.29 & 2.47 & 3.53 & 2.30 & 3.33
 & 2.01 & 3.02 & 1.84 & 2.50 & 2.22 & 3.28 \\
 & + SWA
 & 2.26 & 3.34 & 2.45 & 3.64 & 2.05
 & 3.15 & 2.31 & 3.41 & 2.24 & 3.26
 & 1.75 & 2.79 & 1.70 & 2.45 & 2.09 & 3.15 \\
 & \alg
 & \textbf{2.80} & \textbf{4.02} & \textbf{2.87} & \textbf{4.21} & \textbf{2.52}
 & \textbf{3.78} & \textbf{2.78} & \textbf{4.06} & \textbf{2.95} & \textbf{4.12}
 & \textbf{2.18} & \textbf{3.36} & \textbf{2.13} & \textbf{3.01} & \textbf{2.58} & \textbf{3.79} \\
\bottomrule
\end{tabular}}
\label{tab:main}
\vspace{-10pt}
\end{table*}

%% file: tex/04_experiments.tex
\section{Experimental Evaluation}

\subsection{Experimental Setup}
We evaluate \alg on Qwen3-4B~\cite{yang2025qwen3} and Phi-4-reasoning. All drafters have two layers and use a block size of 7 with chain-based drafting. For training, we follow \citet{cheng2026dspark} and use Open-PerfectBlend, an open-source version of PerfectBlend~\citep{xu2024perfect} containing 1.3 million samples. We replace the original responses with those generated by each target model to improve target alignment. Evaluation covers three task categories: \textbf{Math} (GSM8K; \citealt{cobbe2021training}, MATH; \citealt{lightman2024lets}, AIME25), \textbf{Code} (HumanEval; \citealt{chen2021evaluating}, MBPP; \citealt{austin2021program}, LiveCodeBench; \citealt{jain2025livecodebench}), and \textbf{Chat} (MT-Bench; \citealt{zheng2023judging}). We report average acceptance length ($\tau$) and end-to-end decoding speedup relative to autoregressive decoding, using standard speculative decoding~\citep{leviathan2023fast} at a sampling temperature of 1.0. The detailed experimental setups are in \autoref{app:setup}.

\subsection{Main Results}
\paragraph{General benchmarks.} \autoref{tab:main} summarizes speculative decoding performance across math, code, and chat benchmarks. \alg consistently improves both decoding speedup and acceptance length across Qwen3-4B and Phi-4-reasoning, outperforming DFlash and DSpark across all seven benchmarks. Compared with DSpark, \alg increases average decoding speedup from 1.95$\times$ to 2.15$\times$ on Qwen3-4B and from 2.22$\times$ to 2.58$\times$ on Phi-4-reasoning, corresponding to improvements of 10.0\% and 16.2\%, respectively. Average acceptance length similarly increases from 3.67 to 4.10 and from 3.28 to 3.79. On Phi-4-reasoning, \alg also outperforms the SWA variant, suggesting that recurrent prefix memory effectively complements local attention to improve draft quality and decoding efficiency. These results suggest that the prefix-memory prior improves draft quality sufficiently to yield end-to-end decoding gains despite the additional recurrent computation.

\input{fig/long}
\paragraph{Long-context evaluation.} We further evaluate \alg on three LongBench~\citep{bai2023longbench} tasks, including HotpotQA~\citep{yang-etal-2018-hotpotqa}, QASPER~\citep{dasigi-etal-2021-dataset}, and GovReport~\citep{huang-etal-2021-efficient}, with context lengths ranging from 4K to 32K tokens (Table~\ref{tab:long}). Across all tasks and context lengths, \alg consistently outperforms both full-context and sliding-window attention. Compared with SWA, \alg achieves higher acceptance by preserving historical information through recurrent memory while maintaining bounded context storage. Meanwhile, full-context attention exhibits substantial degradation beyond the 8K training length, whereas \alg maintains robust acceptance with bounded context memory.

\subsection{Ablation Studies}
We further analyze \alg using Phi-4, with additional results provided in \autoref{app:exp}.

\input{fig/component}
\input{fig/layers}
\paragraph{Role of hybrid context modeling.} We examine the complementary roles of recurrent memory and local attention through inference-time ablations (\autoref{tab:intervention}). \emph{Last 512} and \emph{Last feature} restrict the SSM history to recent committed target features, while \emph{Cross-slot off} disables attention among draft positions while preserving attention to recent target features. Restricting recurrent history degrades acceptance on distant evidence, while retaining the recent 512 features largely preserves nearby and late-slot acceptance. In contrast, disabling cross-slot attention impairs nearby and late-slot predictions even with full recurrent history. These complementary effects indicate that recurrent history and intra-block attention serve distinct functional roles during drafting, supporting the proposed separation of long-range context modeling and local interactions under limited capacity.

\paragraph{Effect of drafter capacity.}
We investigate the effect of drafter capacity by varying backbone depth, with approximately matched parameter counts between \alg and full-attention drafters at each depth. As shown in \autoref{fig:capacity}, \alg consistently outperforms full-context attention-only drafters, with the performance gap narrowing as depth increases. Notably, a one-layer \alg is competitive with two-layer full-attention drafters, while its two-layer variant approaches the performance of five-layer counterparts. These results suggest that the prefix-memory prior provides an effective inductive bias for capacity-constrained drafting, with larger benefits in shallow models.

\input{fig/quantization}
\paragraph{Robustness to weight quantization.} To assess the practicality of \alg under tight memory budgets, we quantize the drafter weights to 4 bits using GPTQ~\citep{frantar2023optq}, while keeping the target model unchanged. As shown in \autoref{tab:quantization}, the quantized drafter retains 99.2\% of the BF16 acceptance length on average, while improving decoding speedup by 4.3\%. These results demonstrate that the benefits of the prefix-memory prior are largely preserved under aggressive weight quantization, strengthening the case for \alg under constrained drafter weight budgets.

\subsection{Generality of Prefix-Memory Prior}
\input{fig/ssm}

\paragraph{Mamba-2 vs. Gated DeltaNet-2.}
To examine whether \alg depends on the choice of recurrent operator, we replace Mamba-2 with Gated DeltaNet-2 (GDN-2; \citealt{hatamizadeh2026gated}), while keeping local attention and fusion unchanged. Unlike Mamba-2, GDN-2 employs channel-wise decay and separate erase and write gates, updating its recurrent state as
\begin{align}
&S_i = \bigl(I-k_i(b_i\odot k_i)^\top\bigr)D_iS_{i-1}
      +k_i(w_i\odot v_i)^\top,\nonumber\\
&D_i = \operatorname{Diag}(\alpha_i),
\end{align}
where $b_i$ and $w_i$ control erasure and writing, respectively. The state is updated only from committed target features and remains frozen during drafting. As shown in \autoref{tab:ssm}, both operators achieve comparable acceptance lengths and decoding speedups, with Mamba-2 performing slightly better on average. These results suggest that the benefits of the prefix-memory prior are not specific to Mamba-2.

\input{fig/readout}
\paragraph{DSpark vs. DFlash2.}
To examine the generality of \alg across different proposal mechanisms, we replace the DSpark readout with that of DFlash2~\citep{inco2026dflash2}, including its dynamic local convolutions and pairwise path selector, while keeping the hybrid backbone unchanged. As shown in \autoref{tab:readout}, \alg achieves strong acceptance lengths and decoding speedups with both readouts, with DFlash2 providing modest but consistent improvements across all benchmarks. These results suggest that the prefix-memory prior is compatible with different proposal mechanisms rather than being tied to the DSpark readout.

%% file: fig/long.tex
\begin{table}[t]
\centering
\caption{Average acceptance length ($\tau$) of Phi-4-reasoning on LongBench across context lengths.}
\vspace{-5pt}
\resizebox{1.0\linewidth}{!}{
\begin{tabular}{cccccccccc}
\toprule[0.1em]
\multirow{2}{*}{Context} & \multicolumn{3}{c}{HotpotQA} & \multicolumn{3}{c}{QASPER} & \multicolumn{3}{c}{GovReport} \\ 
\cmidrule(lr){2-4}\cmidrule(lr){5-7}\cmidrule(lr){8-10}
    & Full  & SWA & Hy. & Full  & SWA & Hy. & Full  & SWA & Hy. \\
\midrule
2K--4K  & 2.78 & 2.72 & \textbf{3.02} & 2.61 & 2.45 & \textbf{2.70} & 2.43 & 1.88 & \textbf{2.50} \\
4K--8K  & 3.23 & 2.57 & \textbf{3.25} & 2.52 & 2.31 & \textbf{2.64} & 2.50 & 1.66 & \textbf{2.61} \\
8K--16K  & 1.24 & 2.25 & \textbf{3.11} & 1.79 & 2.12 & \textbf{2.77} & 1.19 & 1.57 & \textbf{2.60} \\
16K--32K  & 1.02 & 2.13 & \textbf{2.72} & 0.97 & 1.81 & \textbf{2.83} & 0.86 & 1.58 & \textbf{2.44} \\
\bottomrule[0.1em]
\end{tabular}}\label{tab:long}
\vspace{-10pt}
\end{table}

%% file: fig/component.tex
\begin{table}[tb]
\centering
\caption{Ablations of recurrent memory and intra-block attention. Nearby/Distance: average acceptance length on synthetic retrieval probes; Late-slot Acc.: conditional acceptance at draft positions 4--7 on natural prefixes.}
\vspace{-5pt}
\resizebox{1.0\linewidth}{!}{
\begin{tabular}{@{}lrrr@{}}
\toprule[0.1em]
Intervention & Nearby ($\uparrow$) & Distance ($\uparrow$) & Late-slot Acc. ($\uparrow$)\\
\midrule
\alg & 6.32 & 5.62 & 0.77 \\
\midrule
\multicolumn{4}{@{}l}{\textit{Recurrent history}} \\
SSM: Last 512 & 6.32 & 3.09 & 0.77 \\
SSM: Last feature & 4.66 & 2.19 & 0.68 \\
\midrule
\multicolumn{4}{@{}l}{\textit{Intra-block interactions}} \\
Cross-slot off & 5.68 & 4.95 & 0.71 \\
Cross-slot off + SSM512 & 5.69 & 2.88 & 0.71 \\
\bottomrule[0.1em]
\end{tabular}}
\label{tab:intervention}
\end{table}

%% file: fig/layers.tex
\begin{figure}[t]
    \centering
    \vspace{-6.5pt}
    \includegraphics[width=1.0\columnwidth]{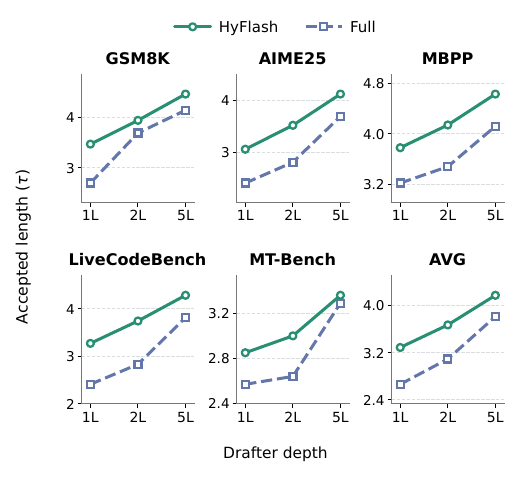}
    \vspace{-20pt}
    \caption{Average acceptance length ($\tau$) of \alg and full-context attention drafters across different depth.}
    \label{fig:capacity}
    \vspace{-10pt}
\end{figure}

%% file: fig/quantization.tex
\begin{table}[t]
\centering
\caption{Effect of 4-bit weight quantization on decoding speedup (Spd.) and average acceptance length ($\tau$) of \alg across math and code benchmarks. Rel. denotes performance relative to BF16.}
\vspace{-5pt}
\resizebox{1.0\linewidth}{!}{
\begin{tabular}{lcccccccc}
\toprule[0.1em]
\multirow{2}{*}{Method} &
\multicolumn{2}{c}{MATH-500} & \multicolumn{2}{c}{AIME25} &
\multicolumn{2}{c}{MBPP} & \multicolumn{2}{c}{LCB} \\
\cmidrule(lr){2-3}\cmidrule(lr){4-5}\cmidrule(lr){6-7}\cmidrule(lr){8-9}
& Spd. & $\tau$ & Spd. & $\tau$ & Spd. & $\tau$ & Spd. & $\tau$ \\
\midrule
BF16    & 2.87 & \textbf{4.21} & 2.52 & \textbf{3.78} & 2.95 & \textbf{4.12} & 2.18 & \textbf{3.36} \\
INT4    & \textbf{2.98} & 4.18 & \textbf{2.61} & 3.72 & \textbf{3.08} & 4.09 & \textbf{2.30} & 3.35 \\
\midrule
$\Delta$ Rel. (\%) & 3.8 & $-$0.7 & 3.6 & $-$1.6 & 4.4 & $-$0.7 & 5.5 & $-$0.7 \\
\bottomrule[0.1em]
\end{tabular}}
\label{tab:quantization}
\end{table}

%% file: fig/ssm.tex
\begin{table}[t]
\centering
\caption{Effect of recurrent operators on decoding speedup (Spd.) and average acceptance length ($\tau$).}
\vspace{-5pt}
\resizebox{1.0\linewidth}{!}{
\begin{tabular}{lcccccccc}
\toprule[0.1em]
\multirow{2}{*}{Method} &
\multicolumn{2}{c}{MATH-500} & \multicolumn{2}{c}{AIME25} &
\multicolumn{2}{c}{MBPP} & \multicolumn{2}{c}{LCB} \\
\cmidrule(lr){2-3}\cmidrule(lr){4-5}\cmidrule(lr){6-7}\cmidrule(lr){8-9}
& Spd. & $\tau$ & Spd. & $\tau$ & Spd. & $\tau$ & Spd. & $\tau$ \\
\midrule
Mamba-2     & \textbf{2.87} & \textbf{4.21} & \textbf{2.52} & \textbf{3.78} & \textbf{2.95} & \textbf{4.12} & 2.18 & 3.36 \\
GDN-2       & 2.85 & 4.14 & 2.47 & 3.59 & 2.76 & 4.01 & \textbf{2.18} & \textbf{3.38} \\
\bottomrule[0.1em]
\end{tabular}}
\label{tab:ssm}
\vspace{-10pt}
\end{table}

%% file: fig/readout.tex
\begin{table}[t]
\centering
\caption{Effect of proposal readout mechanisms on decoding speedup (Spd.) and average acceptance length ($\tau$). Both methods use the same \alg backbone, differing only in their proposal readout mechanisms.}
\vspace{-5pt}
\resizebox{1.0\linewidth}{!}{
\begin{tabular}{lcccccccc}
\toprule[0.1em]
\multirow{2}{*}{Method} &
\multicolumn{2}{c}{MATH-500} & \multicolumn{2}{c}{AIME25} &
\multicolumn{2}{c}{MBPP} & \multicolumn{2}{c}{LCB} \\
\cmidrule(lr){2-3}\cmidrule(lr){4-5}\cmidrule(lr){6-7}\cmidrule(lr){8-9}
& Spd. & $\tau$ & Spd. & $\tau$ & Spd. & $\tau$ & Spd. & $\tau$ \\
\midrule
DSpark      & 2.87 & 4.21 & 2.52 & 3.78 & 2.95 & 4.12 & 2.18 & 3.36 \\
DFlash2     & 2.92 & 4.48 & 2.61 & 3.94 & 3.05 & 4.30 & 2.26 & 3.58 \\
\bottomrule[0.1em]
\end{tabular}}
\label{tab:readout}
\vspace{-10pt}
\end{table}

%% file: tex/05_conclusion.tex
\section{Conclusion}
On-device memory constraints limit drafter capacity and context storage. \alg introduces a prefix-memory prior that combines recurrent modeling of historical context with local attention for recent and intra-block interactions. Capacity scaling and targeted interventions suggest that this inductive bias is particularly effective for shallow parallel drafters, improving acceptance over attention-only baselines while maintaining bounded draft-side context memory.

\clearpage
\section*{Limitations}
Our study focuses on shallow parallel drafters under tight memory constraints. While \alg combines local attention with a fixed-size recurrent state, the latter provides a compressed representation of historical information rather than exact token-level access to the entire prefix. Furthermore, we primarily investigate two-layer drafters with a fixed attention window, leaving alternative capacity allocations and adaptive memory configurations for future exploration.

\section*{Acknowledgments}
We would like to thank Rohith Kukkala, Karthik Vijayan, Sara Abdali, Rhea Sukthanker, and Michael Sapienza for  valuable discussions and support during the course of this work.

%% file: tex/91_related.tex
\section{Related Work}\label{sec:related}
Here, we discuss prior work on speculative decoding, efficient long-context modeling, and memory-efficient drafting. We highlight how these research directions motivate our design and distinguish \alg from existing approaches.

\paragraph{Speculative decoding.} Speculative decoding accelerates generation by verifying tokens proposed by a lightweight drafter~\citep{leviathan2023fast}. While EAGLE~\citep{li2024eagle} and EAGLE-3~\citep{li2025eagle} employ autoregressive drafting, recent methods such as DFlash~\citep{chen2026dflash}, Domino~\citep{huang-etal-2026-domino}, DSpark~\citep{cheng2026dspark}, and DFlash2~\citep{inco2026dflash2} explore parallel drafting with improved inter-token dependencies. In contrast, \alg focuses on efficient context modeling for parallel drafting.

\paragraph{Efficient long-context modeling.} SSMs such as Mamba~\citep{gu2024mamba} and Mamba-2~\citep{dao2024ssm} maintain compact recurrent states for efficient sequence modeling.
Hybrid architectures, including Griffin~\citep{de2024griffin}, Jamba~\citep{lenz2025jamba}, and Hymba~\citep{dong2024hymba}, combine recurrence with attention for efficient language modeling. Building on this principle, \alg adapts hybridization to parallel drafting: committed target features update persistent recurrent memory, while speculative draft positions query the read-only state and interact through local attention over recent context and other draft positions.

\paragraph{Memory-efficient speculative drafting.} MagicDec~\citep{sadhukhan2025magicdec} reduces draft KV-cache overhead, while Mamba Drafters~\citep{choi2025mamba} employ recurrent drafters. More closely related, H-Spec~\citep{jiang2026hspec} combines Mamba and attention for parallel drafting, initializing recurrent states from the target's last-token features and reusing its KV cache. Unlike H-Spec, which initializes recurrent states from the target's last-token features and scans draft positions, \alg maintains a drafter-owned persistent state updated only from committed
target features.

%% file: tex/94_hardware.tex
\input{fig/hardware}
\section{Memory Budget and Synthetic Diagnostic}\label{app:hardware}

\paragraph{Hardware context for memory budget.} Our illustrative 12 GiB inference budget in \autoref{fig:motivation} is motivated by the memory capacities of commercially available on-device AI platforms. As summarized in \autoref{tab:hardware}, representative devices offer 8--32 GB of physical memory, with 16 GB configurations available across consumer and embedded platforms, including Microsoft Surface Laptop ~\citep{microsoft2024surface}, Apple Mac ini~\citep{apple2024mini}, ASUS Vivobook S 15~\citep{asusVivobookS15}, NVIDIA Jetson Orin NX~\citep{nvidiaJetPack62}, and Qualocmm Snapdragon Dev Kit~\citep{qualcommSnapdragonDevKit}. Since physical memory is shared with the operating system and other workloads, the entire capacity cannot be dedicated to inference. We therefore adopt a 12 GiB inference budget for 16 GB-class devices, leaving approximately 2.9 GiB of headroom. Within this budget, target weights, KV cache, runtime overhead, and drafter weights and context must all coexist. 

\paragraph{Synthetic retrieval diagnostic.}
For \autoref{fig:motivation}b, we construct label-copy tasks by inserting eight randomly generated three-digit numbers into a prefix of irrelevant filler text. We fix the prefix length at 6,144 tokens and vary the evidence distance over $\{256, 2048, 5120\}$ tokens. We evaluate 47 jointly filtered prefixes with eight stochastic verification draws per prefix at temperature 1.0. All drafters use identical target-generated prefixes and anchors, and acceptance counts only accepted draft proposals.

%% file: fig/hardware.tex
\begin{table}[t]
\centering
\caption{Representative on-device AI hardware configurations.
RAM denotes total physical memory, while peak AI compute is
reported in tera operations per second (TOPS). Vendor-reported
TOPS may reflect different compute units, numerical precisions,
and sparsity settings and are not directly comparable.
All specifications are taken from publicly available manufacturer
documentation~\citep{microsoft2024surface,apple2024mini,
asusVivobookS15,nvidiaJetPack62,qualcommSnapdragonDevKit}.}
\vspace{-5pt}
\resizebox{1.0\linewidth}{!}{
\begin{tabular}{lcc}
\toprule[0.1em]
\textbf{Platform (Processor)} & \textbf{RAM} & \textbf{Peak AI Compute} \\
\midrule
Microsoft Surface Laptop 7 (X Elite) & 16 GB & 45 TOPS \\
Apple Mac mini (M4) & 16 GB & 38 TOPS \\
ASUS Vivobook S 15 (X Elite) & 16 GB & 45 TOPS \\
NVIDIA Jetson Orin NX & 16 GB & 157 TOPS \\
NVIDIA Jetson Orin Nano Super & 8 GB & 67 TOPS \\
Qualcomm Snapdragon Dev Kit (X Elite) & 32 GB & 45 TOPS \\
\bottomrule[0.1em]
\end{tabular}}
\label{tab:hardware}
\vspace{-10pt}
\end{table}

%% file: tex/92_experimental_setup.tex
\section{Detailed Experimental Settings}\label{app:setup}
In this section, we provide additional details on the drafter configurations, training procedures, and evaluation protocols used in our experiments.

\paragraph{Drafter configuration.} We use two-layer drafters with a block size of $K=7$ for both Qwen3-4B and Phi-4-reasoning. We build \alg on DSpark~\citep{cheng2026dspark}, retaining its target-feature conditioning, parallel block drafting, and rank-256 Markov proposal head. We extract hidden features from five uniformly spaced target layers, spanning the second to the third-to-last layer. Adapted target features are provided to each drafter layer, while only draft-slot representations propagate between layers. We replace the attention mixer with our hybrid design combining SWA ($W=512$) and Mamba-2~\citep{dao2024ssm}. We retain the target model's hidden size and feed-forward configuration while splitting the attention projection width between local attention and a Mamba-2 recurrent branch, keeping the overall parameter count approximately matched to the attention-only baselines. For Phi-4-reasoning, local attention uses 20 query heads and 5 KV heads with a head dimension of 128, while the recurrent branch uses 20 heads with a head dimension and state size of 128. We apply the same parameter-matching design to Qwen3-4B. The target remains frozen, while the feature adapter, drafter backbone, and proposal head are trained.

\paragraph{Baseline configuration.} We compare against DFlash~\citep{chen2026dflash}, DSpark~\citep{cheng2026dspark}, and its SWA variant ($W=512$). All methods use the same target model, proposal length, and speculative verification protocol.

\paragraph{Data preprocessing.} Training conversations are tokenized using the corresponding target tokenizer and chat template, with a maximum sequence length of 8,192 tokens. We sample up to 512 valid assistant-token positions per conversation as drafting anchors, each conditioned only on preceding target features. Multiple blocks are processed in a single drafter forward pass. During training, we causally scan the adapted target features and extract the recurrent state immediately before each sampled anchor. All speculative positions within a draft block query this shared prefix state without updating it.

\paragraph{Optimization.} We use AdamW~\citep{loshchilov2018decoupled} with a peak learning rate of $6\times10^{-4}$, $\beta_1=0.9$, $\beta_2=0.95$, and weight decay of $0.1$. We apply cosine decay with 4\% linear warmup and gradient clipping at $1.0$. Training uses BF16 weights, FP32 optimizer states, a global batch size of 512 conversations, and 10 epochs.

\paragraph{Training objective.} Following \citet{cheng2026dspark}, we combine cross-entropy (CE) loss for ground-truth token prediction with total variation (TV) loss for matching draft and target distributions:
\begin{align}
    &\mathcal{L}_{\mathrm{CE}} = -\sum_{k=1}^{K} w_k \log p_k^d(x_k^*), \nonumber \\
    &\mathcal{L}_{\mathrm{TV}} = \sum_{k=1}^{K} w_k \|p_k^d - p_k^t\|_1, \nonumber \\
    &\mathcal{L} = \alpha_{\mathrm{CE}} \mathcal{L}_{\mathrm{CE}} + \alpha_{\mathrm{TV}} \mathcal{L}_{\mathrm{TV}},
\end{align}
where $p_k^d$ and $p_k^t$ denote the draft and target distributions, respectively. We use $\alpha_{\mathrm{CE}}=0.1$, $\alpha_{\mathrm{TV}}=0.9$, and position weights $w_k=\exp(-(k-1)/4)$~\citep{chen2026dflash}.

\paragraph{Inference configuration.} Unless otherwise specified, we use batch-one decoding without top-$p$~\citep{Holtzman2020The} or top-$k$ truncation. The proposal head samples tokens sequentially from a single backbone pass. For \alg, recurrent states are updated only with newly committed target features after verification and remain read-only during drafting. Rejected proposals do not modify the states.

\paragraph{Inference hardware and runtime.} We measure batch-one decoding throughput on an NVIDIA RTX PRO 6000 Blackwell GPU using BF16 target and drafter weights. End-to-end speedup is measured against autoregressive decoding on the same hardware.

\paragraph{Model size and memory accounting.} For Phi-4-reasoning, the two-layer \alg drafter contains 865.5M parameters, compared with 864.1M for the full-attention and SWA baselines. For Qwen3-4B, the corresponding counts are 303.3M and 312.4M. These counts exclude the input embedding and output vocabulary head, which are reused from the frozen target. The resulting BF16 weight footprints of \alg are 1,650.8 MiB and 578.5 MiB, respectively, with GPTQ 4-bit footprints of 465.9 MiB and 205.6 MiB (estimated for Qwen3-4B). In the Phi-4 configuration, local KV storage and FP32 recurrent states each require approximately 2.5 MiB, yielding a constant 5 MiB logical context footprint matching the SWA baseline. Both methods physically evict out-of-window KV entries. These estimates exclude target KV and temporary buffers, and do not represent measured peak memory.

\paragraph{Metric computation.} For general benchmarks, average acceptance length ($\tau$) counts newly generated tokens per verification round, including eligible bonus tokens but excluding the initial anchor. Decoding speedup is the ratio of generated-token throughput to autoregressive decoding, including drafting, verification, and recurrent-state updates. For frozen-prefix evaluations, including synthetic retrieval~(\autoref{fig:motivation}) and LongBench~(\autoref{tab:long}), we instead report the average number of accepted draft proposals at shared target-generated prefixes, excluding correction and bonus tokens. For aggregate results, we report the geometric mean of decoding speedups and the arithmetic mean of acceptance lengths across benchmarks.

%% file: tex/93_additional.tex
\section{Additional Experimental Discussion}\label{app:exp}
\input{fig/tree}
\input{fig/block}
\input{fig/sample}
\input{fig/context}
\paragraph{Tree-based Drafting.} We evaluate \alg under fixed and adaptive tree drafting with an identical verification budget of 29 nodes. The fixed strategy uses four depth-7 branches, while the adaptive strategy follows the confidence-guided tree construction of BASTION~\citep{oh2026bastion}. As shown in \autoref{tab:tree}, \alg consistently improves both acceptance length and decoding speedup over DSpark under both strategies. These results suggest that the benefits of the prefix-memory prior extend beyond chain-based drafting to different tree construction strategies.

\paragraph{Sensitivity to draft block size.} We vary the inference-time draft block size over $k\in\{3,5,7,9\}$ while using the same drafters trained with $K=7$. As shown in \autoref{tab:draft_length}, \alg consistently outperforms attention-only baselines in acceptance length and decoding speedup across different block sizes. Notably, \alg maintains its advantage at $k=9$, beyond the training block size, suggesting that the prefix-memory prior remains effective for longer draft blocks.

\paragraph{Sensitivity to sampling parameters.} We evaluate \alg under greedy ($T=0$), moderate ($T=0.6$, top-$p=0.95$), and untruncated ($T=1$, top-$p=1$) sampling, using the same trained drafter without modification. As shown in \autoref{tab:sampling}, greedy decoding achieves the highest acceptance length and speedup, while increasing sampling stochasticity reduces both metrics across all benchmarks. Nevertheless, \alg maintains decoding speedups under all sampling configurations, demonstrating its applicability across different sampling distributions.

\paragraph{Sensitivity to context length.} We evaluate \alg under 4K and 16K context lengths while keeping the drafter architecture and training configuration unchanged. As shown in \autoref{tab:context}, acceptance length increases at 16K on MATH-500 and LCB but decreases on AIME25 and MBPP, while decoding speedup declines across all four benchmarks. As further examined in the latency analysis below, this speedup degradation is primarily associated with the increasing cost of full-attention target verification rather than drafter computation.

\input{fig/time}
\paragraph{Latency analysis.} To investigate the reduced decoding speedup at longer contexts, we profile the latency of speculative decoding at 4K and 16K. As shown in \autoref{fig:latency}, target verification latency on AIME25 nearly doubles from 44.76 to 87.80 ms, while draft-backbone latency remains nearly constant (2.95 vs. 3.02 ms). Unlike autoregressive decoding, which processes one query position per forward pass, speculative verification processes eight positions, incurring greater attention overhead as the context grows. Meanwhile, explicit KV-cache eviction and incremental recurrent-state updates keep the drafter computation nearly independent of context length. These results indicate that target verification becomes the dominant bottleneck at longer contexts.

%% file: fig/tree.tex
\begin{table}[t]
\centering
\caption{Effect of tree drafting strategies on decoding speedup (Spd.) and average acceptance length ($\tau$). Fixed uses width 4 and depth 7, while Adaptive follows BASTION~\citep{oh2026bastion}.}
\vspace{-5pt}
\resizebox{1.0\linewidth}{!}{
\begin{tabular}{lcccccccc}
\toprule[0.1em]
\multirow{2}{*}{Method} &
\multicolumn{4}{c}{AIME25} &
\multicolumn{4}{c}{LCB} \\
\cmidrule(lr){2-5}\cmidrule(lr){6-9}
&
\multicolumn{2}{c}{Fixed} & \multicolumn{2}{c}{Adaptive} &
\multicolumn{2}{c}{Fixed} & \multicolumn{2}{c}{Adaptive} \\
\cmidrule(lr){2-3}\cmidrule(lr){4-5}\cmidrule(lr){6-7}\cmidrule(lr){8-9}
& Spd. & $\tau$ & Spd. & $\tau$ & Spd. & $\tau$ & Spd. & $\tau$ \\
\midrule
DSpark        & 2.15 & 3.42 & 2.40 & 3.94 & 1.82 & 2.97 & 2.03 & 3.42 \\
\alg          & \textbf{2.52} & \textbf{3.81} & \textbf{2.74} & \textbf{4.29} & \textbf{2.17} & \textbf{3.44} & \textbf{2.41} & \textbf{3.94} \\
\bottomrule[0.1em]
\end{tabular}}
\label{tab:tree}
\end{table}

%% file: fig/block.tex
\begin{table}[t]
\centering
\caption{Sensitivity to inference-time draft length. Average acceptance length ($\tau$) and decoding speedup for $k\in\{3,5,7,9\}$, using drafters trained with $K=7$.}
\vspace{-5pt}
\resizebox{1.0\linewidth}{!}{
\begin{tabular}{lcccccccc}
\toprule[0.1em]
\multirow{2}{*}{Method} &
\multicolumn{2}{c}{$k=3$} &
\multicolumn{2}{c}{$k=5$} &
\multicolumn{2}{c}{$k=7$} &
\multicolumn{2}{c}{$k=9$} \\
\cmidrule(lr){2-3}\cmidrule(lr){4-5}\cmidrule(lr){6-7}\cmidrule(lr){8-9}
& Spd. & $\tau$ & Spd. & $\tau$ & Spd. & $\tau$ & Spd. & $\tau$ \\
\midrule
DSpark  & 1.88 & 2.73 & 2.12 & 3.10 & 2.22 & 3.28 & 2.18 & 3.27 \\
DSpark + SWA & 1.55 & 2.28 & 1.93 & 2.89 & 2.09 & 3.15 & 2.03 & 3.04 \\
\alg & \textbf{1.95} & \textbf{2.89} & \textbf{2.45} & \textbf{3.68} & \textbf{2.58} & \textbf{3.79} & \textbf{2.62} & \textbf{3.98} \\
\bottomrule[0.1em]
\end{tabular}}
\label{tab:draft_length}
\vspace{-10pt}
\end{table}

%% file: fig/sample.tex
\begin{table}[t]
\centering
\caption{Sensitivity to sampling parameters on decoding speedup (Spd.) and average acceptance length ($\tau$).}
\vspace{-5pt}
\resizebox{1.0\linewidth}{!}{
\begin{tabular}{lcccccccc}
\toprule[0.1em]
\multirow{2}{*}{Method} &
\multicolumn{2}{c}{MATH-500} & \multicolumn{2}{c}{AIME25} &
\multicolumn{2}{c}{MBPP} & \multicolumn{2}{c}{LCB} \\
\cmidrule(lr){2-3}\cmidrule(lr){4-5}\cmidrule(lr){6-7}\cmidrule(lr){8-9}
& Spd. & $\tau$ & Spd. & $\tau$ & Spd. & $\tau$ & Spd. & $\tau$ \\
\midrule
Greedy      & 3.35 & 4.86 & 2.90 & 4.29 & 4.09 & 5.77 & 3.13 & 4.69 \\
Moderate     & 3.22 & 4.75 & 2.79 & 4.25 & 3.00 & 4.39 & 2.54 & 3.97 \\
Untruncated     & 2.87 & 4.21 & 2.52 & 3.78 & 2.95 & 4.12 & 2.18 & 3.36 \\
\bottomrule[0.1em]
\end{tabular}}
\label{tab:sampling}
\end{table}

%% file: fig/context.tex
\begin{table}[t]
\centering
\caption{Sensitivity to context length on decoding speedup (Spd.) and average acceptance length ($\tau$).}
\vspace{-5pt}
\resizebox{1.0\linewidth}{!}{
\begin{tabular}{lcccccccc}
\toprule[0.1em]
\multirow{2}{*}{Method} &
\multicolumn{2}{c}{MATH-500} & \multicolumn{2}{c}{AIME25} &
\multicolumn{2}{c}{MBPP} & \multicolumn{2}{c}{LCB} \\
\cmidrule(lr){2-3}\cmidrule(lr){4-5}\cmidrule(lr){6-7}\cmidrule(lr){8-9}
& Spd. & $\tau$ & Spd. & $\tau$ & Spd. & $\tau$ & Spd. & $\tau$ \\
\midrule
4K      & 2.87 & 4.21 & 2.52 & 3.78 & 2.95 & 4.12 & 2.18 & 3.36 \\
16K     & 2.00 & 5.46 & 1.41 & 2.95 & 1.64 & 3.69 & 1.78 & 4.15 \\
\bottomrule[0.1em]
\end{tabular}}
\label{tab:context}
\vspace{-10pt}
\end{table}

%% file: fig/time.tex
\begin{figure}[t]
    \centering
    \includegraphics[width=1.0\columnwidth]{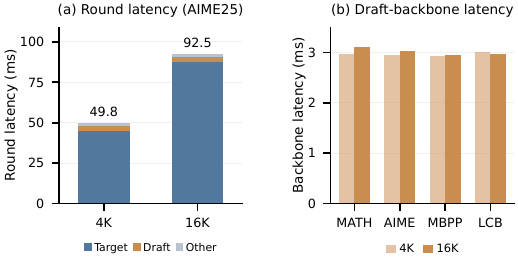}
    \vspace{-20pt}
    \caption{Latency analysis at 4K and 16K context lengths, excluding prefill. (a) Per-round latency breakdown on AIME25. (b) Draft-backbone latency across four benchmarks.}
    \label{fig:latency}
    \vspace{-10pt}
\end{figure}